\documentclass{article}

\usepackage[final, nonatbib]{neurips_2022}

\usepackage[utf8]{inputenc} 
\usepackage[T1]{fontenc}    
\usepackage{hyperref}       
\usepackage{url}            

\usepackage{booktabs}       
\usepackage{amsfonts}       
\usepackage{nicefrac}       
\usepackage{microtype}      
\usepackage{xcolor}         
\usepackage{todonotes}
\usepackage[title]{appendix}

\title{Desiderata for next generation of ML model serving}

\author{%
  Sherif Akoush \thanks{Equal contribution} \ \thanks{Seldon Technologies} \\
  \texttt{sa@seldon.io} \\
  \And
  Andrei Paleyes \footnotemark[1] \ \footnotemark[2] \ \thanks{Department of Computer Science and Technology, University of Cambridge}\\
  \texttt{ap2169@cam.ac.uk} \\
  \And
  Arnaud Van Looveren \footnotemark[2]\\
  \texttt{avl@seldon.io} \\
  \And
  Clive Cox \footnotemark[2]\\
  \texttt{cc@seldon.io} \\
}

\begin{document}

\maketitle

\begin{abstract}
Inference is a significant part of ML software infrastructure. Despite the variety of inference frameworks available, the field as a whole can be considered in its early days. This position paper puts forth a range of important qualities that next generation of inference platforms should be aiming for. We present our rationale for the importance of each quality, and discuss ways to achieve it in practice. We propose to focus on data-centricity as the overarching design pattern which enables smarter ML system deployment and operation at scale.
\end{abstract}

\section{Introduction}

Model inference has become an important part of modern machine learning (ML) infrastructure. Various sources estimate up to 90\% of ML compute resources are used for inference tasks \cite{deepspeedinference,barr2019amazon,leopold2019aws,nn_carbon_google,nn_carbon_fb}. To answer a growing demand for inference infrastructure, a number of model serving platforms appeared on the market~\cite{bentoml, kserve, ray_serve,seldon_core}. In addition, cloud providers are also offering services that simplify model serving for their users~\cite{aws,google}.

While the existing ML model serving frameworks answer some of the initial challenges of ML deployment, we believe there are a number of important properties such frameworks need to have to ensure a seamless model deployment experience. For instance, effective monitoring and explainability of an entire inference pipeline is an open research direction. Efficient use of infrastructure while optimising for important metrics of energy and environmental footprint is missing. Seamless ML deployment to different targets such as edge devices is required. While ML serving frameworks are taking steps to provide some of these features, we think that taking into consideration all desired aspects of the system is challenging \cite{paleyes2022challenges}, requires leveraging different architecture patterns, but in general will lead to better solutions developed by the community.

In this position paper we present a set of desired features for ML deployment --- a blueprint of the next generation ML model serving frameworks. We discuss motivations for each feature and provide initial pointers on ways to achieve them. We hope to bring the community and practitioners together to discuss these challenges, find ideal solutions, and shape the future of ML serving.

\section{Desiderata}
In this section we present nine qualities that are important for ML serving. We advocate that designing the system with data-centricity~\cite{datacentric-2,datacentric-1} as the highest priority enables these features.

\subsection{Inference pipelines as dataflow graphs}\label{dataflow-pipelines}
\begin{figure}
    \centering
    \includegraphics[width=\textwidth]{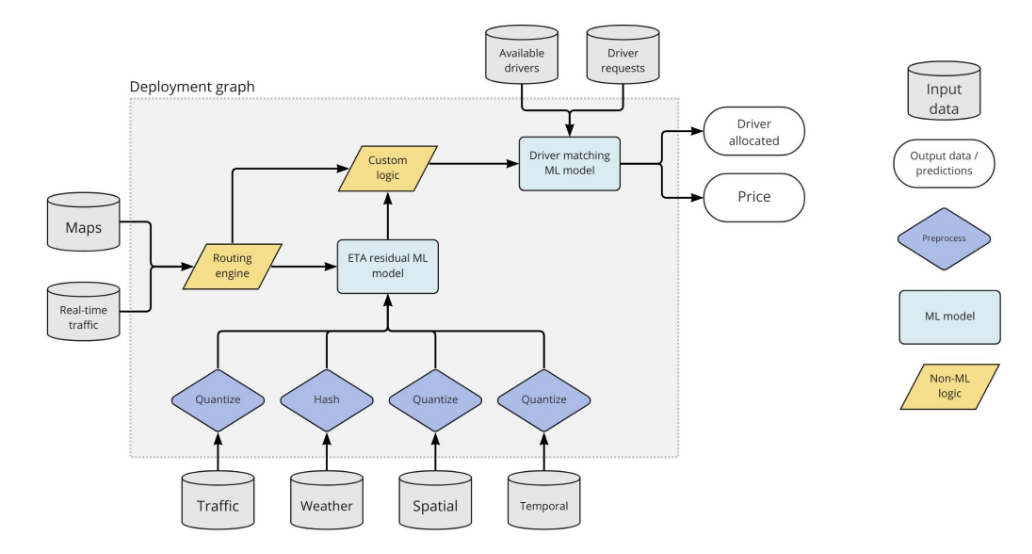}
    \caption{An inference pipeline of a ride-sharing service, motivated by an example from Uber \cite{hu2022deepreta}. It ingests multiple data sources, contains several business logic operations and ML models, and produces several outputs.}
    \label{figure:inference_pipeline}
\end{figure}

Model inference is a complex data processing pipeline. It can include input and output data transformations, multiple ML models, monitoring components, custom business logic, and so on. An example of a complex inference graph is shown in Figure~\ref{figure:inference_pipeline}. It is imperative to have a clear view of the flow of data through the entire pipeline, both for its developers and users. Runtime access to any intermediate data in the pipeline allows for better experimentation, troubleshooting and monitoring experience, all of which are discussed later in this paper.

There are two possible ways for a model serving platform to facilitate access to a dataflow graph at runtime. It can be discovered post-hoc, for example with distributed tracing systems \cite{jaeger,berg2021snicket}. This approach is applicable to platforms implemented with service oriented approaches, such as microservices. While a popular paradigm for software system design, service orientation might be ill-suited for ML inference, as its control flow nature poorly reflects data-centricity of the inference process. Alternatively, inference pipelines can be built with a dataflow-first approach, such as the flow-based programming (FBP) paradigm, as is already the case for ML training pipelines \cite{cai2016tfdebugger,mahapatra2022flow}. FBP provides access to the dataflow graph naturally \cite{paleyes2022empirical}, and thus might be a more appropriate choice for implementation of inference graphs.

\subsection{Pipeline component abstractions}

To allow for the construction of complex data pipelines as shown in Figure~\ref{figure:inference_pipeline}, simple but flexible architectural building blocks are required. ML models usually run on dedicated specialised servers. The models will have generally been created by data scientists while the final serving infrastructure is usually handled by a separate dedicated operations teams. The definition of models and servers for inference should be kept separate to allow for their distinct creation and for the possibility of model sharing across servers.

Given a set of models, many data pipelines may be built which share them at inference time. Consequently, a pipeline abstraction should be a higher level concept that defines the flow of data between the functional steps and how that data is joined and split as needed. Teams should be able to tap into any data source, consume any output stream or extend the inference pipeline for extra processing.

\subsection{Support for synchronous and asynchronous scenarios} \label{sync-async}
Modern ML inference services should natively support two modes of operation: synchronous and asynchronous. Synchronous inference is the traditional mode of doing inference, also known as request-driven batch processing. In this mode a user makes a request with a batch of input data points to the platform, and receives a response with corresponding predictions. For that interaction to happen the inference platform has to provide an API endpoint suitable for accepting prediction requests \cite{pmlr-v50-azureml15}. Asynchronous inference allows for a different interaction pattern, where the input data is arriving and the predictions are produced continuously. Such behaviour can be achieved with data streams, and is often referred to as stream processing \cite{shahrivari2014beyond}.

With real-time ML gaining attention because of its data availability, flexibility and ability to scale \cite{huyen2020realtimeml}, asynchronous use cases will be encountered more often. Stream processing possesses a range of qualities which make it better suited for real-time analytics \cite{wingerath2016real}, such as the ability to adapt to variable workloads \cite{das2014adaptive,lohrmann2015elastic}. At the same time, reports estimate around 77\% of software development teams are working with service-based architectures \cite{oreilly2020microservices}, which means it is easier for many users to include batch processing APIs in their software setup. Remarkably, hybrid approaches are now being proposed, in an attempt to provide convenient interfaces for both synchronous and asynchronous interaction modes \cite{carbone2015apache,dissanayake2017cloud,fino2021rstream}.

\subsection{Seamless deployment to Cloud and Edge}
ML models are trained in the cloud using a big fleet of powerful machines and potentially deployed to many different hardware for inference. For example an image recognition model can be deployed to cameras for online object detection, and the same model can be hosted in a data centre for background feed processing. Even within the same data centre the same ML model might be deployed to different machine specs with or without hardware accelerators.

One of the challenges is to optimise the model performance according to the target hardware. There are existing tools that can help automatically generate target-specific artefacts employing techniques such as ML model quantisation, operator fusion and network pruning~\cite{tvm, deepspeedinference, quantise, quantise2, googleinference}. However it is a complex multidimensional optimisation problem that should take into consideration cost, latency and throughput targets.

Therefore the ML serving framework should take care of setting up the infrastructure as required, optimising the inference service accordingly. ML serving frameworks shall be able to provide the same set of features to the users, regardless of where models are deployed.

\subsection{Flexible experimentation}\label{experimentaion}

ML is an iterative task where new components are updated over time to better harness the data that is currently flowing through them. Therefore, an inference system needs flexible and simple ways to test updates both to individual atomic models as well as entire data pipelines. Furthermore, as the number of data pipelines increases the task of validating experiments manually becomes more onerous. In recent years the use of progressive rollouts for new models where clear service level agreements (SLAs) and acceptance criteria are defined and the data traffic is managed automatically between challenger and champion models is becoming more popular with tools on the market~\cite{argorollouts, flagger, iter8}. These techniques need to be extended to handle entire data pipeline rollouts in an efficient manner. Managing experiments at scale also requires smarter monitoring which we describe next.

\subsection{Monitoring and explainability}\label{rca}
The key responsibility of monitoring is to identify abnormal situations in the inference system and allow fast and accurate root-cause analysis (RCA). A good monitoring solution for ML systems needs to be able to flexibly allow RCA for any connected set of components in the inference graph. Access to the dataflow graph at runtime, as discussed in Section \ref{dataflow-pipelines}, provides the necessary intermediate data streams and runtime graph traversal operations to achieve such flexibility.

A particular example of such a flexible monitoring feature is context-aware drift detection \cite{cobb2022context}, where a data drift observed downstream can be analysed jointly with some of the upstream data used as a context. Another example is recursive attribution \cite{singal2021flow}, where an observed behaviour is linked to the most likely cause, and this procedure is repeated recursively until the user input is reached.

Operations and auditing teams need to be able to explain on demand any part of a pipeline for technical, regulatory or business reasons \cite{klaise2020monitoring}. When data snapshots itself are not sufficient to derive such explanations, inference systems should allow to dynamically add ML interpretation models.

\subsection{Continual and active learning}

ML model performance in production degrades over time. To address this challenge the first step is to detect performance issues via in-depth RCA as described in Section~\ref{rca}. The next step is to trigger (re)training of the problematic ML model on new data, compare metrics between the existing and candidate model, and automatically rollout the new version to production. Therefore ML serving frameworks should integrate well with ML training tools, exposing data from production that can improve the (re)training process. An extension is to support active learning~\cite{activelearning} techniques in which there is a tight integration between data acquisition, incremental (re)training and inference. 

\subsection{Data privacy and compliance}
With more ML-driven solutions used to automate sensitive decision making, ML applications become governed by various legislatures, such as GDPR in Europe, PIPEDA in Canada or APPI in Japan. Users that run their workloads on model inference platforms have to protect privacy of the data passing through the pipeline, prove their compliance with relevant regulations, and provide explanations of decisions made by their models.

Consequently, it is important for ML inference frameworks to aid compliance and provide privacy guarantees. Access to the complete dataflow graph can be a reasonable way to achieve these requirements, as it can facilitate privacy assessment \cite{tang2022assessing} and enable ``compliance by construction'' \cite{schwarzkopf2019position}. In particular, framework users need to be able to answer questions such as \textit{What are the assumptions behind the observed output? What are the main input factors which contributed the most to the output? What minimal changes to the input can change the output?}. Collectively such meta information about data is known as data provenance~\cite{carata2014primer}, and the ability to systematically answer these questions is a critical mechanism for assuring compliance.

Literature describes multiple attacks against deployed models, such as model inversion \cite{veale2018algorithms} and model stealing \cite{tramer2016stealing}, mostly targeted at recovering parts of private training datasets and model parameters. Defence methods against such attacks are being actively explored \cite{oliynyk2022know}, and should be considered as a part of any modern model inference platform.

\subsection{Resource efficiency}
\textbf{Detailed cost and energy metrics.} Doing inference over the lifetime of an ML model is energy hungry. Inference can reach up to 90\% of the energy consumed during the ML model lifecycle~\cite{ai_compute, nn_carbon_google,nn_carbon_fb}. While this adds costs to organisations, as ML becomes integrated in our daily lives we should strive to have a holistic accounting of ML energy use that incorporates all related tasks~\cite{hopper}. The community is already taking steps to address some of these challenges and account for energy and $CO_2$ emissions of ML training tasks \cite{code_carbon,nn_carbon_google, nn_carbon_fb}. These techniques should be extended to ML serving.

\textbf{Multimodel serving with overcommit of resources.} We advocate for multimodel serving pattern where one ML inference server hosts multiple models at the same time. This reduces overheads required to deploy a large number of models while making it simpler to operate the system at scale (check Appendix~\ref{mms_example} for more details). Multimodel serving is already a feature provided by some inference servers and frameworks~\cite{kserve, mlserver, triton}. Moreover, in many cases demand patterns allow for further optimisation such as overcommit of resources. This means that an ML system could register more models than what can be served by the provisioned infrastructure. The system should be able to swap models dynamically according to usage without adding significant latency overheads to inference requests. A complementary approach is autoscaling of resources (e.g. replicas of inference nodes) according to load.

\section{Conclusion}
With this position paper we hope to reinforce the importance of research and development on ML inference systems and encourage a debate on the desired requirements for the next generation of tools to successfully deploy and manage powerful predictive pipelines.

\bibliographystyle{abbrv}
\bibliography{references}


\clearpage

\begin{appendices}

\section{Multimodel vs single model serving ~\label{mms_example}}
For serving ML models in production, a standard pattern is to package up the ML model inside a container image which then gets deployed and managed by a service orchestration framework (e.g. Kubernetes). A slight variation of this pattern is to have the ML model persisted separately (e.g. in an object store) and the orchestration framework fetching and injecting the model artifact inside the container runtime during startup. While this pattern works well for organisations in the case of deploying a couple of models, it does not scale well as there is a one-to-one mapping between a deployed container and an ML model being served. With the requirement to deploy many thousands of models in production there is going to be a lot of extra resource overhead to keep these containers running in the system.

To illustrate this concept with a working example there is a current Kubernetes limitation on the number of pods per node, which is 110 pods per node \cite{k8s_large_cluster}.
To deploy 20,000 single pod ML models we would need then ~200 nodes. The smallest node in Google Cloud is \texttt{e2-micro} (1 GB memory) and therefore the total system would require at least ~200 GB of provisioned memory. In fact the memory requirement is likely to be far greater as it is not possible to have 100s of model inference server pods on a small node. With multimodel serving however the memory footprint of the system is expected to be one order of magnitude less by design as resources are shared at the model level.

Multimodel serving has also additional benefits. It allows for better CPU/GPU sharing. It does not suffer from the issue of cold start, where with each ML model to deploy we have to download the container image before starting it --- this is usually in the order of tens of minutes. Multimodel serving also reduces the risk of allocating new cloud resource (e.g. GPU) on-demand as model inference servers are long-lived by design.
\end{appendices}


\end{document}